\def\year{2024}\relax
\documentclass[letterpaper]{article} 
\usepackage{aaai24}  
\usepackage{times}  
\usepackage{helvet}  
\usepackage{courier}  
\usepackage[hyphens]{url}  
\usepackage{graphicx} 
\usepackage{natbib}  
\usepackage{caption} 
\usepackage{multirow}
\usepackage{xcolor}
\usepackage{verbatim}
\usepackage{array}
\usepackage{tabularx}
\usepackage{subcaption}
\DeclareCaptionStyle{ruled}{labelfont=normalfont,labelsep=colon,strut=off} 
\usepackage{amsmath,amssymb}

\usepackage{algorithm}
\usepackage{algorithmic}
\usepackage[nameinlink,capitalise]{cleveref}

\usepackage{soul}
\setstcolor{red}
\usepackage{xspace}
\usepackage[colorinlistoftodos,prependcaption,textsize=tiny,textwidth=15mm]{todonotes}
\definecolor{navy}{rgb}{0.1, 0.1, 0.8}
\definecolor{gray}{rgb}{0.4, 0.4, 0.4}
\definecolor{ruby}{rgb}{0.8, 0.2, 0.4}
\definecolor{olive}{rgb}{0.1, 0.5, 0.1}
\definecolor{plotviolet}{RGB}{136, 25, 203}
\definecolor{plotgreen}{RGB}{103, 202, 77}

\newcommand{\eat}[1]{}

\usepackage{newfloat}
\usepackage{listings}
\floatstyle{ruled}
\newfloat{listing}{tb}{lst}{}
\floatname{listing}{Listing}
\title{Misinformation is not about Bad Facts: An Analysis of the Production and Consumption of Fringe Content}
\author{
    JooYoung Lee,\textsuperscript{\rm 1}
    Emily Booth,\textsuperscript{\rm 1}
    Hany Farid,\textsuperscript{\rm 2}
    Marian-Andrei Rizoiu\textsuperscript{\rm 1}
}
\affiliations {
\textsuperscript{\rm 1} University of Technology Sydney \\
\textsuperscript{\rm 2} School of Information, University of California, Berkeley \\
\{Jooyoung.Lee, Emily.Booth, Marian-Andrei.Rizoiu\}@uts.edu.au, hfarid@berkeley.edu
}

\begin{document}
\maketitle
\begin{abstract}

What if misinformation is not an information problem at all? To understand the role of news publishers in potentially unintentionally propagating misinformation, we examine how far-right and fringe online groups share and leverage established legacy news media articles to advance their narratives.  Our findings suggest that online fringe ideologies spread through the use of content that is consensus-based and ``factually correct''. We found that Australian news publishers with both moderate and far-right political leanings contain comparable levels of information completeness and quality; and furthermore, that far-right Twitter users often share from moderate sources. However, a stark difference emerges when we consider two additional factors: 1) the narrow topic selection of articles by far-right users, suggesting that they cherry pick only news articles that engage with their preexisting worldviews and specific topics of concern, and 2) the difference between moderate and far-right publishers when we examine the writing style of their articles. Furthermore, we can identify users prone to sharing misinformation based on their communication style. These findings have important implications for countering online misinformation, as they highlight the powerful role that personal biases towards specific topics and publishers' writing styles have in amplifying fringe ideologies online.
\end{abstract}

%
\section{Introduction}
\label{sec:introduction}

Misinformation has historically been understood as an information or factual accuracy issue, where concerns arise when inaccurate narratives emerge despite the existence of a social or expert consensus on the topic. This has led to a focus on interventions that seek to remedy inaccuracies, like fact-checking, presuming that people will willingly change their perspective on an issue when presented with an alternative, authoritative form of information. These approaches, however, overlook the human side of the problem, such as the everyday anxieties and grievances that can motivate belief in misinformation~\citep{booth2024conspiracy}. The ubiquity of digital technologies and social media means the reach and consequences of misinformation are increasingly amplified~\citep{Attwell2021,Shearer:2021vb,kydd2021decline}. 

Technological interventions by social media platforms have thus far proved to be limited in controlling the spread of misinformation~\citep{johns2024labelling}.  As a result, it follows that we must turn our attention to the heart of the issue: that humans have uncertainties and worries, and seek solace through answers. Thus, when considering misinformation online, it is crucial to consider what is being shared not as an abstract issue of factual inaccuracy, but in its context: who is sharing it, who are the likely consumers, and how is it being shared to reach audiences effectively.

To understand the role of news publishers in potentially unintentionally propagating misinformation, we examine how far-right and fringe online groups share and leverage established legacy news media articles to advance their narratives. We also demonstrate that far-right and moderate news articles in our sample differ not in the information completeness of the articles, but in their writing style, thus providing a new perspective on misinformation as an issue of style. From this, we consider how style might be leveraged by moderate and consensus-based news sources to better counter the influence of online misinformation.

\textbf{Research Questions.} We use content- and style-based measures to show that misinformation, like any other content, can be styled to target particular online cohorts. 
In particular, we answer the following research questions (RQs):
\begin{itemize}
    \item \textbf{RQ1:} \textit{What role does the style and content of news media play in enabling the spread of misinformation?}
    \item \textbf{RQ2} 
    {\it Can we differentiate fringe groups based on linguistic styles?}
    \item \textbf{RQ3}
    {\it Can we differentiate contents from producers and consumers based on linguistic styles? } 
\end{itemize}
%

%
%
\begin{figure}[tbp]
    \includegraphics[scale=0.32]{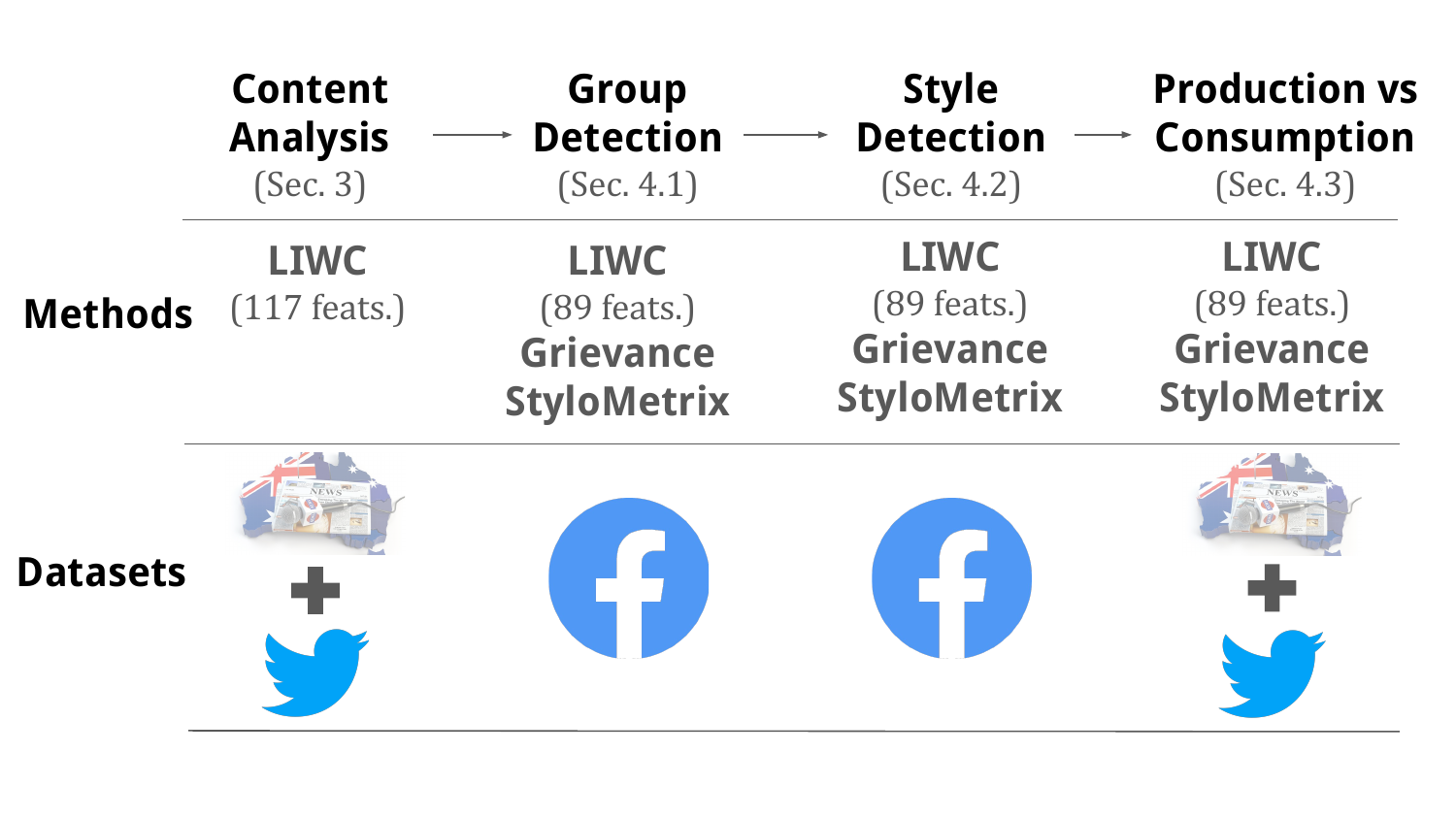}
    \centering
    \caption{Summary of the methods: we start by analysing contents of Australian news articles and sharing patterns of those articles in Twitter environment. Then we use linguistic styles -- not the actual article contents -- measured with \textsc{LIWC}, \textsc{Grievance} dictionary and \textsc{StyloMetrix} to identify extreme groups in Facebook. Next, we identify text styles employed by the extreme groups and classify them. Finally, we distinguish misinformation production and consumption using styles.}
    \label{fig:introduction}
    \end{figure}

The graphic in~\Cref{fig:introduction} depicts the steps of our analysis, the datasets used and the methods. We begin by analyzing the content completeness of news articles based on their political ideology. We compare 1) publishers of different ideologies and 2) production and consumption by far-right users. Based on the findings that content completeness of news articles does not significantly differ based on publishers' ideology, we leverage the text style (instead of its content) to investigate differences and distinguish between publishers of different ideologies and fringe online communities. Lastly, we show that far-right production and consumption can be distinguished based on the employed writing style.   
We expand on the details of our contributions below. 

We conducted a comprehensive analysis of Australian news media consumption, focusing on several key aspects. Firstly, we utilized the Trust Index~\cite{tde} -- a metric measuring information completeness -- to compare news articles from legacy news media publishers to those with distinct ideological stances. Secondly, we examined the differences in writing style across these sources. Additionally, we delved into the sharing patterns of news articles by highly partisan online users. This differs from much of the prior work that has started from the assumption that misinformation is an information problem, and studied the connection between misinformation and the publishers' political leaning~\citep{calvillo2020political,das2022rethinking}. 

Our analysis revealed a distinct difference between what articles were published by far-right news sources and what articles were shared by the extreme (here, also far-right) ideological cohort in our sample, particularly in terms of linguistic patterns. However, we notably found less distinction between far-right and moderate articles. Even in cases where differences were statistically significant, the effect sizes were small, indicating that the sharing of extreme content is less tied to the source's perceived ideological stance and more about how users interpret the value of an article as a tool to advance their own worldview. This is to say that users will individually cherry pick articles to share on the basis of their own views, regardless of what source published it.

This leads us to the conclusion that misinformation is not strictly a matter of factually-inaccurate content being spread by far-right sources, but that mostly-accurate content can be selectively consumed to fuel an existing fringe worldview. To explore this further, we investigate how such groups express their worldviews through different linguistic styles.


This study uses stylistic metrics from established dictionaries (such as \textsc{LIWC} and the \textsc{Grievance} dictionary) to construct stylistic classifiers.
Using these stylistic classifiers, we can accurately distinguish consumers who may be vulnerable to misinformation (like anti-vaccination and far-right sentiment) from regular online social media users.
We further show that we can detect extreme ideological users' writing styles, as categorized by our manual labeling. 
Lastly, we show that we can successfully distinguish far-right sources from far-right consumers using a stylistic classifier.

The summary of our contributions are as follows: 
\begin{enumerate}
    \item We provide an analysis that comparatively examines the ideological landscape of Australian news and how users selectively share news to further their own ends regardless of the source ({\bf RQ1}).
    \item We propose a classifier to identify linguistic styles within extreme online groups as well as styles that are commonly used in the online misinformation space ({\bf RQ2}).
    \item We show that the production and consumption of misinformation exhibit different patterns that can be differentiated by styles ({\bf RQ3}).
\end{enumerate}

\section{Datasets and Measures}
\label{dataset}

We present the datasets and measures used in this work. 

\subsection{News and social media datasets}

\subsubsection{Australian news by Google News.}
To evaluate the potential role of Australian news publications in facilitating misinformation dissemination, we collected Australian news articles sourced from Google News via The Daily Edit (TDE) platform.
TDE aggregates news articles from Google News, encompassing 14 distinct topics: `Climate Change,' `Sport,' `Human Migration,' `World,' `Finance,' `Technology,' `Taiwan,' `Top Stories,' `Entertainment,' `Australia,' `Business,' `Health,' `Science,' and `China.' 
This is the exhaustive list of topics available when the region is set to Australia (AU) in Google News. 
The published period for the news articles spans from \textsf{March 1, 2014}, to \textsf{January 12, 2023}. 
The dataset contains $214,754$ articles published by $9,929$ news publishers.

TDE computes a \emph{Trust Index} score for each article as follows.
Articles about the same event are grouped into \emph{stories}.
Articles are represented as sequences of sentences; all similar sentences within a story are clustered together, and clusters are interpreted as \emph{details}. 
Consequently, each article has a set of supporting details (common narrative elements across multiple sources in a story).
TDE computes an article's \emph{Trust Index} based on the percentage of details it covers from a story -- the article's informational completeness.
%

\subsubsection{The political leaning of publishers.}
We use an external media bias dataset from \textit{allsides}\footnote{https://www.allsides.com}, which assesses the political leaning of $473$ news publishers on a five-point scale, ranging from extreme- and moderate-left to center to moderate- and extreme-right. These media bias ratings represent the average viewpoint of individuals across the political spectrum rather than the perspective of any single individual or group~\cite{park2021digital, newman2021reuters}. We consolidate the extreme- and moderate-left categories into a single `left' class and merge the extreme- and moderate-right categories into a `right' class. This results in three political leaning classes: left, center, and right. Later, we also examine articles from publishers  belonging to the `extreme-right' range by \textit{allsides}.

\begin{table*}[t]
    \caption{Number of news articles by topic and political leaning.}
    \label{tab:numart}
    \tiny
    \centering
    \resizebox{\textwidth}{!}{
      \begin{tabular}{c|cllcccccc}
           & \multicolumn{3}{c}{Top Stories}   & Australia     & World         & Technology   & Sport           & Entertainment  & Health       \\ \hline
      L     & \multicolumn{3}{c}{4264 ($39\%$)} & 2346 ($33\%$) & 2288 ($38\%$) & 671 ($46\%$) & 1270 ($41\%$)   & 1165 ($35\%$)  & 396 ($39\%$) \\
      C     & \multicolumn{3}{c}{2783 ($26\%$)} & 2193 ($31\%$) & 1894 ($31\%$) & 539 ($37\%$) & 609 ($20\%$)    & 993 ($30\%$)   & 355 ($33\%$) \\
      R     & \multicolumn{3}{c}{3768 ($35\%$)} & 2504 ($36\%$) & 1876 ($31\%$) & 259 ($18\%$) & 1243 ($40\%$)   & 1179 ($35\%$)  & 257 ($25\%$) \\
      Total & \multicolumn{3}{c}{10815}         & 7043          & 6058          & 1469         & 3122            & 3337           & 1008         \\ \hline \\
            & \multicolumn{3}{c}{China}         & Business      & Science       & Finance      & Human migration & Climate change & Taiwan       \\ \hline
      L     & \multicolumn{3}{c}{908 ($41\%$)}  & 720 ($36\%$)  & 319 ($33\%$)  & 435 ($43\%$) & 442 ($41\%$)    & 318 ($46\%$)   & 48 ($38\%$)  \\
      C     & \multicolumn{3}{c}{846 ($39\%$)}  & 668 ($33\%$)  & 513 ($53\%$)  & 382 ($38\%$) & 436 ($41\%$)    & 270 ($39\%$)   & 44 ($35\%$)  \\
      R     & \multicolumn{3}{c}{435 ($20\%$)}  & 621 ($31\%$)  & 128 ($13\%$)  & 200 ($20\%$) & 198 ($18\%$)    & 105 ($15\%$)   & 34 ($27\%$)  \\
      Total & \multicolumn{3}{c}{2189}          & 2009          & 960           & 1017         & 1076            & 693            & 126         
      \end{tabular}
  }
  \end{table*}

\subsubsection{Far-right Twitter users.}
We use a manually curated list containing $1,496$ Australian far-right coded Twitter users~\cite{sociologyFriends} and collect their $3,665,809$ most recent tweets using the Twitter API (at most $3,200$ tweets for each user). 
The collected tweets span from \textsf{February 22, 2009} to \textsf{December 9, 2022}. These tweets contained $1,827,162$ {\it urls} consisting of $11,643$ articles in Google News dataset (\cref{sec:newsmedia}). We expand the articles by linking to the publishers present in either the Google News or {\it allsides} dataset which results in $215,242$ news articles (\cref{styleovercontent}). We scraped the content of each article using \textsf{urllib.parse} Python library\footnote{ \url{https://docs.python.org/3/library/urllib.parse.html}}. 
We successfully downloaded the texts of $155,669$ articles at the end because the rest of the links were not available.


\subsubsection{Facebook groups.}
We manually assemble two lists of Facebook pages for specific ideologies, namely Australian \texttt{Antivax} and \texttt{Far-right} pages. Australian \texttt{Far-right} has $14$ Facebook pages and Australian \texttt{Antivax} has six Facebook pages. 
Using the CrowdTangle API, we collect all posts from the Facebook pages between \textsf{January 23, 2021} and \textsf{February 24, 2023}. We collect $6,017$ posts from the \texttt{Far right} pages and $2,969$ posts from the \texttt{Antivax} pages. 

\subsection{Linguistic Measurements of Style}
\label{subsec:linguistic-measurements}

We utilize three linguistic metrics that quantify linguistic attributes within text. 

\subsubsection{\textsc{LIWC}} is one of the most widely used text analysis tools in psychology, which has recently been adopted by computational social scientists to draw insights into human behavior through computational methods~\cite{chung2018we}. 
This tool captures words relating to content (e.g., death, religion) and function (e.g., conjunctions, articles). 
LIWC (version 2022)\footnote{https://www.liwc.app/} has $117$ categories in total. 
However, to capture the extreme groups' stylistic differences without contamination from content, we removed content-related features (such as {\it religion}, {\it family} and others -- see \cref{liwc_list}), yielding a total of $89$ style-related categories. 

\subsubsection{\textsc{Grievance}} is a psycholinguistic dictionary to capture language use in the context of grievance-fueled violence threat assessment~\cite{van2021grievance}. 
Grudge has been shown as the critical ingredient that distinguishes militant extremist mindset from social conservatism~\cite{stankov2021social}; grudge (alongside confusion) is also one of the ingredients of misinformation consumption~\cite{booth2024conspiracy}.
We include \textsc{Grievance} dictionary
due to its ability to extract violence and threat-specific words. 

\subsubsection{\textsc{StyloMetrix}} is a grammar-related statistical representation of text.
This tool allows for representing a text sample of any length with a linguistic vector of a fixed size~\cite{okulskastyles} offering 
several preferable characteristics over well-known contextual embedding, such as BERT. 
First, \textsc{StyloMetrix} vectors encode entire documents, resolving the issue of varying text lengths. 
This could help when combining texts from multiple platforms, such as Facebook and Twitter. 
Second, \textsc{StyloMetrix} vectors aim to encode the entire sample's stylistic structure, not the words' meanings. 

%
\section{The Content: News Production and Sharing}
\label{sec:newsmedia}

Here, we answer RQ1 and demonstrate that there is only a marginal difference in the information completeness of left- and right-leaning publishers, with the left-leaning publishers publishing slightly more articles. 
Furthermore, there are few differences in linguistic patterns between far-right and moderate publishers. However, we see a marked difference when comparing what far-right publishers \textit{produce} compared to what far-right users actually \textit{share} online.

\subsection{Coverage and Trust of publishers}
\label{subsec:coverage-trust}

\subsubsection{Coverage by topic.}
We extracted the $40,922$ Google News articles from publishers with identified stances based on the media bias data (note that not all publishers from the Google News dataset are present in the \textit{allsides} dataset, and we exclude any missing publishers).
\cref{tab:numart} shows the number of articles from the stance-identified publishers for each topic.
Across all topics, except `Entertainment,' we observe a higher number of left-leaning articles than right-leaning. Particularly noteworthy is the disparity in percentages within the `Climate change' and `Technology' topics, where the count of right-leaning articles is substantially lower than that of left-leaning articles. This indicates that the Australian news media is generally perceived to be more left-leaning.

%
\begin{table*}[t]
    \caption{Trust Index statistics of news articles for all topics. The $p$-values were derived from $t$-tests conducted on the L and R groups. The effect size values are calculated using Cohen's $d$. Statistically significant ($p<0.05$) results are highlighted.}
    \label{tab:alltopic}
    \small
    \centering
    \resizebox{\textwidth}{!}{
      \begin{tabular}{llccccccccccccccccccccc}
        \multicolumn{2}{c}{}             & \multicolumn{3}{c|}{Top Stories} & \multicolumn{3}{c|}{Australia} & \multicolumn{3}{c|}{World}      & \multicolumn{3}{c|}{Technology} & \multicolumn{3}{c|}{Sport}          & \multicolumn{3}{c|}{Entertainment} & \multicolumn{3}{c}{Health}                                                                                                                                                                                                                     \\
        \multirow{4}{*}{Trust Index}     & \multicolumn{1}{l|}{}            & $\mu$                          & $\sigma$                        & \multicolumn{1}{c|}{N}          & $\mu$                               & $\sigma$                           & \multicolumn{1}{c|}{N}     & $\mu$ & $\sigma$ & \multicolumn{1}{c|}{N}    & $\mu$ & $\sigma$ & \multicolumn{1}{c|}{N}   & $\mu$ & $\sigma$ & \multicolumn{1}{c|}{N}    & $\mu$ & $\sigma$ & \multicolumn{1}{c|}{N}    & $\mu$ & $\sigma$ & N   \\ \cline{2-23}
                                         & \multicolumn{1}{c|}{L}           & 0.60                           & 0.20                            & \multicolumn{1}{c|}{4264}       & 0.63                                & 0.21                               & \multicolumn{1}{c|}{2346}  & 0.60  & 0.20     & \multicolumn{1}{c|}{2288} & 0.58  & 0.19     & \multicolumn{1}{c|}{671} & 0.62  & 0.20     & \multicolumn{1}{c|}{1270} & 0.61  & 0.19     & \multicolumn{1}{c|}{1165} & 0.61  & 0.20     & 396 \\
                                         & \multicolumn{1}{c|}{C}           & 0.59                           & 0.20                            & \multicolumn{1}{c|}{2783}       & 0.61                                & 0.21                               & \multicolumn{1}{c|}{2193}  & 0.57  & 0.21     & \multicolumn{1}{c|}{1894} & 0.55  & 0.20     & \multicolumn{1}{c|}{539} & 0.55  & 0.20     & \multicolumn{1}{c|}{609}  & 0.63  & 0.20     & \multicolumn{1}{c|}{993}  & 0.58  & 0.21     & 355 \\
                                         & \multicolumn{1}{c|}{R}           & 0.54                           & 0.20                            & \multicolumn{1}{c|}{3768}       & 0.60                                & 0.21                               & \multicolumn{1}{c|}{2504}  & 0.53  & 0.20     & \multicolumn{1}{c|}{1876} & 0.51  & 0.19     & \multicolumn{1}{c|}{259} & 0.55  & 0.20     & \multicolumn{1}{c|}{1243} & 0.60  & 0.19     & \multicolumn{1}{c|}{1179} & 0.59  & 0.20     & 257 \\ \hline
        \multicolumn{2}{l|}{Effect size} & \multicolumn{3}{c|}{{\bf 0.31}}  & \multicolumn{3}{c|}{{\bf 0.11}}      & \multicolumn{3}{c|}{{\bf 0.36}} & \multicolumn{3}{c|}{{\bf 0.36}} & \multicolumn{3}{c|}{{\bf 0.35}}     & \multicolumn{3}{c|}{0.03}          & \multicolumn{3}{c}{0.10}                                                                                                                                                                                                                       \\ \hline
        \multicolumn{2}{l|}{p value}     & \multicolumn{3}{c|}{{\bf 0.0000}}      & \multicolumn{3}{c|}{{\bf 0.0001}}    & \multicolumn{3}{c|}{{\bf 0.0000}}     & \multicolumn{3}{c|}{{\bf 0.0000}}     & \multicolumn{3}{c|}{{\bf 0.0000}}         & \multicolumn{3}{c|}{0.5287}        & \multicolumn{3}{c}{0.2218}                                                                                                                                                                                                                     \\
        \hline
        \\
  
        \multicolumn{2}{l}{}             & \multicolumn{3}{c}{China}        & \multicolumn{3}{c}{Business}   & \multicolumn{3}{c}{Science}     & \multicolumn{3}{c}{Finance}     & \multicolumn{3}{c}{Human migration} & \multicolumn{3}{c}{Climate change} & \multicolumn{3}{c}{Taiwan}                                                                                                                                                                                                                     \\
        \multirow{4}{*}{Trust Index}     & \multicolumn{1}{l|}{}            & $\mu$                          & $\sigma$                        & \multicolumn{1}{c|}{N}          & $\mu$                               & $\sigma$                           & \multicolumn{1}{c|}{N}     & $\mu$ & $\sigma$ & \multicolumn{1}{c|}{N}    & $\mu$ & $\sigma$ & \multicolumn{1}{c|}{N}   & $\mu$ & $\sigma$ & \multicolumn{1}{c|}{N}    & $\mu$ & $\sigma$ & \multicolumn{1}{c|}{N}    & $\mu$ & $\sigma$ & N   \\ \cline{2-23}
                                         & \multicolumn{1}{l|}{L}           & 0.60                           & 0.20                            & \multicolumn{1}{c|}{908}        & 0.63                                & 0.19                               & \multicolumn{1}{c|}{720}   & 0.61  & 0.19     & \multicolumn{1}{c|}{319}  & 0.58  & 0.18     & \multicolumn{1}{c|}{435} & 0.58  & 0.21     & \multicolumn{1}{c|}{442}  & 0.57  & 0.18     & \multicolumn{1}{c|}{318}  & 0.66  & 0.20     & 48  \\
                                         & \multicolumn{1}{l|}{C}           & 0.58                           & 0.20                            & \multicolumn{1}{c|}{846}        & 0.60                                & 0.20                               & \multicolumn{1}{c|}{668}   & 0.62  & 0.19     & \multicolumn{1}{c|}{513}  & 0.54  & 0.18     & \multicolumn{1}{c|}{382} & 0.54  & 0.20     & \multicolumn{1}{c|}{436}  & 0.57  & 0.20     & \multicolumn{1}{c|}{270}  & 0.55  & 0.19     & 44  \\
                                         & \multicolumn{1}{l|}{R}           & 0.52                           & 0.20                            & \multicolumn{1}{c|}{435}        & 0.61                                & 0.19                               & \multicolumn{1}{c|}{621}   & 0.60  & 0.19     & \multicolumn{1}{c|}{128}  & 0.52  & 0.18     & \multicolumn{1}{c|}{200} & 0.51  & 0.21     & \multicolumn{1}{c|}{198}  & 0.53  & 0.18     & \multicolumn{1}{c|}{105}  & 0.57  & 0.20     & 34  \\ \hline
        \multicolumn{2}{l|}{Effect size} & \multicolumn{3}{c|}{{\bf 0.38}}  & \multicolumn{3}{c|}{0.08}      & \multicolumn{3}{c|}{0.08}       & \multicolumn{3}{c|}{{\bf 0.33}} & \multicolumn{3}{c|}{{\bf 0.31}}     & \multicolumn{3}{c|}{0.22}          & \multicolumn{3}{c}{0.40}                                                                                                                                                                                                                       \\ \hline
        \multicolumn{2}{l|}{p value}     & \multicolumn{3}{c|}{{\bf 0.0000}}      & \multicolumn{3}{c|}{0.1414}    & \multicolumn{3}{c|}{0.4423}     & \multicolumn{3}{c|}{{\bf 0.0001}}     & \multicolumn{3}{c|}{{\bf 0.0003}}         & \multicolumn{3}{c|}{0.0695}        & \multicolumn{3}{c}{0.0793}
      \end{tabular}}
  \end{table*}

\subsubsection{Informational completeness by political leaning.}
We conduct a comparative analysis of the Trust Index for articles from left-, center-, and right-leaning publishers. 
For each topic and political group, \cref{tab:alltopic} provides the mean ($\mu$), standard deviation ($\sigma$), and the number of articles ($N$). 
We assess whether there is a significant difference between Trust Index of left- and right-leaning news articles using independent samples $t$-tests.
We report the test \textit{p}-value, and we emphasize statistically significant results ($p < 0.05$).
We also quantify the effect size for each test using Cohen's \textit{d}~\cite{cohen2013statistical}.
Generally, a \textit{d} value of $0.2$ indicates a small effect size and a value of $0.5$ is considered a medium effect.

There are several observations from \cref{tab:alltopic}.
First, the mean Trust Index for articles from left-leaning publishers (L in \cref{tab:alltopic}) consistently surpasses that of the right-leaning, regardless of the topic's statistical significance.
This indicates that, in general, left-leaning articles tend to be more informationally complete.
The standard deviations of the Trust Index are nearly identical across all groups and topics.
Second, several topics such as `World', `China', `Technology', `Finance' and `Human Migration' achieve statistical significance but only moderate effect size.
`Taiwan' exhibits the largest effect size among all topics; 
however, it does not reach statistical significance due to the limited number of articles in the `Taiwan' category (refer to \cref{tab:numart}). 
Conversely, `Australia' shows a statistically significant difference, albeit with only a small effect size (Cohen's $d=0.11$). 

\subsubsection{Conclusion.}
While the distinction between left- and right-leaning publishers' information completeness is statistically significant for specific topics, the practical impact of this difference is marginal in the real world, given the small or non-existent effect size. This is unexpected, as we might assume fringe news outlets to consistently include misinformation or factual inaccuracies.



\subsection{Is the content from far-right publishers different from the moderates?}
\label{subsec:far-right-publishers}

Here, we demonstrate there is insufficient evidence using LIWC to show that Google News articles' content and writing style differ based on the political leaning of the publishers.
More specifically, we find this conclusion is consistent with far-right publishers (`extreme-right' per {\it allsides} classification) compared to all the others that we group here as ``moderates".

\subsubsection{Setup.}

We qualitatively inspected randomly selected articles and identified four specific topics -- `Top Stories,' `Australia,' `Finance,' and `Climate Change' -- that exhibited interesting content contrasts.
\cref{tab:art_text_LIWC} shows the number of articles from far-right and moderate publishers for each category.
We gather the content of each article using the \textsf{beautifulsoup} Python library\footnote{\textsc{beautifulsoup}: \url{https://pypi.org/project/beautifulsoup4/}}.
Next, we use LIWC to analyze the text of each article and perform {\it t-tests} to find whether articles from far-right publishers show statistically significant differences from moderates.
To present significant results, we use the Bonferroni correction, i.e., $p < \alpha/117$ where $\alpha = 0.05$ (note: there are $117$ categories in LIWC, see \cref{subsec:linguistic-measurements}).



%
\begin{table*}[]
    \caption{
        Comparative analysis of the article's content published by far-right publishers and the rest (far-left, left, center and right). 
        We only report the LIWC features that show statistically significant differences. Large effect sizes are boldfaced.
    }
    \small
    \centering
    \resizebox{\textwidth}{!}{
        \begin{tabular}{lcrrllllll}
            \multicolumn{1}{c}{\textbf{}}         &                                        & \multicolumn{2}{c}{\textbf{Top Stories}} & \multicolumn{2}{c}{\textbf{Australia}} & \multicolumn{2}{c}{\textbf{Finance}} & \multicolumn{2}{c}{\textbf{Climate change}}                                                                                                                                         \\ \hline
                                                  & \multirow{2}{*}{\textbf{Num articles}} & \multicolumn{1}{c}{Moderate}             & \multicolumn{1}{c}{Far-right}          & \multicolumn{1}{c}{Moderate}         & \multicolumn{1}{c}{Far-right}               & \multicolumn{1}{c}{Moderate}      & \multicolumn{1}{c}{Far-right} & \multicolumn{1}{c}{Moderate}      & \multicolumn{1}{c}{Far-right} \\
                                                  &                                        & 9459                                     & 148                                    & \multicolumn{1}{r}{6880}             & \multicolumn{1}{r}{46}                      & \multicolumn{1}{r}{1073}          & \multicolumn{1}{r}{13}        & \multicolumn{1}{r}{881}           & \multicolumn{1}{r}{12}        \\ \hline
            \hline
            \textbf{Significant categories}       & \multicolumn{1}{l}{}                   & \multicolumn{1}{l}{}                     & \multicolumn{1}{l}{}                   &                                      &                                             &                                   &                               &                                   &                               \\ \hline
            \textsc{Authentic}                    & $\mu$                                  & 34.79                                    & 28.37                                  &                                      &                                             &                                   &                               &                                   &                               \\
                                                  & $\sigma$                               & 23.47                                    & 20.50                                  &                                      &                                             &                                   &                               &                                   &                               \\
                                                  & Effect size                            & \multicolumn{2}{c}{0.27}                 &                                        &                                      &                                             &                                   &                               &                                                                   \\ \hline
            \textsc{Words Per Sentence}                          & $\mu$                                  & 23.77                                    & 20.64                                  & \multicolumn{1}{r}{23.50}            & \multicolumn{1}{r}{21.11}                   &                                   &                               &                                   &                               \\
                                                  & $\sigma$                               & 9.96                                     & 4.09                                   & \multicolumn{1}{r}{6.90}             & \multicolumn{1}{r}{4.18}                    &                                   &                               &                                   &                               \\
                                                  & Effect size                            & \multicolumn{2}{c}{0.32}                 & \multicolumn{2}{c}{0.35}               &                                      &                                             &                                   &                                                                                                   \\ \hline
            \multicolumn{1}{r}{\textsc{pronoun}}  & $\mu$                                  & 7.47                                     & 8.73                                   &                                      &                                             &                                   &                               &                                   &                               \\
                                                  & $\sigma$                               & 4.17                                     & 3.70                                   &                                      &                                             &                                   &                               &                                   &                               \\
                                                  & Effect size                            & \multicolumn{2}{c}{0.30}                 &                                        &                                      &                                             &                                   &                               &                                                                   \\ \hline
            \multicolumn{1}{r}{\textsc{ppron}}    & $\mu$                                  & 4.48                                     & 5.86                                   &                                      &                                             &                                   &                               &                                   &                               \\
                                                  & $\sigma$                               & 3.19                                     & 3.19                                   &                                      &                                             &                                   &                               &                                   &                               \\
                                                  & Effect size                            & \multicolumn{2}{c}{0.34}                 &                                        &                                      &                                             &                                   &                               &                                                                   \\ \hline
            \multicolumn{1}{r}{\textsc{you}}      & $\mu$                                  & 0.34                                     & 0.69                                   &                                      &                                             &                                   &                               &                                   &                               \\
                                                  & $\sigma$                               & 1.11                                     & 0.91                                   &                                      &                                             &                                   &                               &                                   &                               \\
                                                  & Effect size                            & \multicolumn{2}{c}{0.32}                 &                                        &                                      &                                             &                                   &                               &                                                                   \\ \hline
            \multicolumn{1}{r}{\textsc{shehe}}    & $\mu$                                  & 1.77                                     & 2.36                                   &                                      &                                             &                                   &                               &                                   &                               \\
                                                  & $\sigma$                               & 1.94                                     & 1.93                                   &                                      &                                             &                                   &                               &                                   &                               \\
                                                  & Effect size                            & \multicolumn{2}{c}{0.30}                 &                                        &                                      &                                             &                                   &                               &                                                                   \\ \hline
            \multicolumn{1}{r}{\textsc{tentat}}   & $\mu$                                  & 1.17                                     & 1.42                                   &                                      &                                             &                                   &                               &                                   &                               \\
                                                  & $\sigma$                               & 1.13                                     & 0.81                                   &                                      &                                             &                                   &                               &                                   &                               \\
                                                  & Effect size                            & \multicolumn{2}{c}{0.22}                 &                                        &                                      &                                             &                                   &                               &                                                                   \\ \hline
            \textsc{Social}                       & $\mu$                                  & 10.67                                    & 12.11                                  &                                      &                                             &                                   &                               &                                   &                               \\
                                                  & $\sigma$                               & 4.89                                     & 4.57                                   &                                      &                                             &                                   &                               &                                   &                               \\
                                                  & Effect size                            & \multicolumn{2}{c}{0.29}                 &                                        &                                      &                                             &                                   &                               &                                                                   \\ \cline{2-10}
            \multicolumn{1}{r}{\textsc{polite}}   & $\mu$                                  & 0.38                                     & 0.19                                   &                                      &                                             &                                   &                               &                                   &                               \\
                                                  & $\sigma$                               & 0.78                                     & 0.31                                   &                                      &                                             &                                   &                               &                                   &                               \\
                                                  & Effect size                            & \multicolumn{2}{c}{0.24}                 &                                        &                                      &                                             &                                   &                               &                                                                   \\ \cline{2-10}
            \multicolumn{1}{r}{\textsc{comm}}     & $\mu$                                  & 2.01                                     & 2.54                                   &                                      &                                             &                                   &                               &                                   &                               \\
                                                  & $\sigma$                               & 1.61                                     & 1.30                                   &                                      &                                             &                                   &                               &                                   &                               \\
                                                  & Effect size                            & \multicolumn{2}{c}{0.29}                 &                                        &                                      &                                             &                                   &                               &                                                                   \\ \cline{2-10}
            \multicolumn{1}{r}{\textsc{male}}     & $\mu$                                  & \multicolumn{1}{l}{}                     & \multicolumn{1}{l}{}                   &                                      &                                             & \multicolumn{1}{r}{0.51}          & \multicolumn{1}{r}{1.63}      &                                   &                               \\
                                                  & $\sigma$                               & \multicolumn{1}{l}{}                     & \multicolumn{1}{l}{}                   &                                      &                                             & \multicolumn{1}{r}{1.11}          & \multicolumn{1}{r}{1.32}      &                                   &                               \\
                                                  & Effect size                            & \multicolumn{1}{l}{}                     & \multicolumn{1}{l}{}                   &                                      &                                             & \multicolumn{2}{c}{\textbf{1.01}} &                               &                                                                   \\ \hline
            \textsc{Lifestyle}                    & $\mu$                                  & 4.70                                     & 3.60                                   & \multicolumn{1}{r}{5.06}             & \multicolumn{1}{r}{2.81}                    &                                   &                               &                                   &                               \\
                                                  & $\sigma$                               & 3.18                                     & 1.71                                   & \multicolumn{1}{r}{3.60}             & \multicolumn{1}{r}{1.71}                    &                                   &                               &                                   &                               \\
                                                  & Effect size                            & \multicolumn{2}{c}{0.35}                 & \multicolumn{2}{c}{0.63}               &                                      &                                             &                                   &                                                                                                   \\ \cline{2-10}
            \multicolumn{1}{r}{\textsc{work}}     & $\mu$                                  & 2.59                                     & 2.08                                   & \multicolumn{1}{r}{3.11}             & \multicolumn{1}{r}{1.51}                    &                                   &                               &                                   &                               \\
                                                  & $\sigma$                               & 2.36                                     & 1.42                                   & \multicolumn{1}{r}{2.72}             & \multicolumn{1}{r}{1.22}                    &                                   &                               &                                   &                               \\
                                                  & Effect size                            & \multicolumn{2}{c}{0.22}                 & \multicolumn{2}{c}{0.59}               &                                      &                                             &                                   &                                                                                                   \\ \cline{2-10}
            \multicolumn{1}{r}{\textsc{money}}    & $\mu$                                  & 1.10                                     & 0.44                                   & \multicolumn{1}{r}{1.48}             & \multicolumn{1}{r}{0.41}                    &                                   &                               &                                   &                               \\
                                                  & $\sigma$                               & 2.09                                     & 1.10                                   & \multicolumn{1}{r}{2.14}             & \multicolumn{1}{r}{0.70}                    &                                   &                               &                                   &                               \\
                                                  & Effect size                            & \multicolumn{2}{c}{0.33}                 & \multicolumn{2}{c}{0.50}               &                                      &                                             &                                   &                                                                                                   \\ \hline
            \textsc{AllPunc}                      & $\mu$                                  & 16.45                                    & 17.98                                  &                                      &                                             &                                   &                               &                                   &                               \\
                                                  & $\sigma$                               & 4.66                                     & 4.36                                   &                                      &                                             &                                   &                               &                                   &                               \\
                                                  & Effect size                            & \multicolumn{2}{c}{0.33}                 &                                        &                                      &                                             &                                   &                               &                                                                   \\ \hline
            \multicolumn{1}{r}{\textsc{time}}     & $\mu$                                  & \multicolumn{1}{l}{}                     & \multicolumn{1}{l}{}                   & \multicolumn{1}{r}{5.08}             & \multicolumn{1}{r}{3.87}                    &                                   &                               &                                   &                               \\
                                                  & $\sigma$                               & \multicolumn{1}{l}{}                     & \multicolumn{1}{l}{}                   & \multicolumn{1}{r}{2.67}             & \multicolumn{1}{r}{1.64}                    &                                   &                               &                                   &                               \\
                                                  & Effect size                            & \multicolumn{2}{l}{}                     & \multicolumn{2}{c}{0.45}               &                                      &                                             &                                   &                                                                                                   \\ \hline
            \multicolumn{1}{r}{\textsc{emo\_neg}} & $\mu$                                  & \multicolumn{1}{l}{}                     & \multicolumn{1}{l}{}                   &                                      &                                             &                                   &                               & \multicolumn{1}{r}{0.26}          & \multicolumn{1}{r}{0.65}      \\
                                                  & $\sigma$                               & \multicolumn{1}{l}{}                     & \multicolumn{1}{l}{}                   &                                      &                                             &                                   &                               & \multicolumn{1}{r}{0.37}          & \multicolumn{1}{r}{0.41}      \\
                                                  & Effect size                            & \multicolumn{1}{l}{}                     & \multicolumn{1}{l}{}                   &                                      &                                             &                                   &                               & \multicolumn{2}{c}{\textbf{1.03}}                                 \\ \hline
        \end{tabular}
    }
    \label{tab:art_text_LIWC}
\end{table*}

\subsubsection{Results.}
\cref{tab:art_text_LIWC} presents the 21 pairs (LIWC category, article topic) for which the difference between the far-right publishers and the moderates is statistically significant -- 14 LIWC categories for `Top stories,' 5 for `Australia' and one for each of `Finance' and `Climate change.'
However, most of these have low ($d \sim 0.2$) or moderate ($d \sim 0.5$) effect sizes; only two such topic-category pairs show large effect sizes ($d \gtrsim 0.8$).
In `Finance', far-right publishers use significantly more words from the \textsc{male} category -- containing $230$ words such as `he,' `his,' `him' or `man' ($d = 1.01$).
The second is `Climate change', with far-right articles using statistically significantly more words in the \textsc{Negative Emotions} category -- $618$ words such as `bad,' `hate,' `hurt,' `worry,' `fear' ($d = 1.03$). 

\subsubsection{Conclusion.}
The linguistic signals captured by LIWC alone are insufficient for distinguishing extreme (far-right) publishers from moderate ones.
Corroborated with the conclusion from \cref{subsec:coverage-trust}, this indicates little difference in the content and topic of what publishers of different political leanings produce overall.   

\subsection{Far-right articles: content production vs consumption}
\label{pvss}
%
\begin{table*}[]
    \caption{
        Analyzing far-right information consumption (articles shared by far-right Twitter users) with information production (far-right and moderate news media).
        We report select LIWC features that show statistically significant differences against the far-right Twitter-shared articles. 
        Values in \textcolor{gray!60}{grey color} are non-statistically significant. 
        The Cohen's $d$ values are calculated for each topic (`Top Stories', `Australia', `Finance' and `Climate change') between the Google News and Twitter-shared groups.}
    \label{tab:farrighttext}
    \resizebox{\textwidth}{!}{
   \begin{tabular}{llc|ccc|ccc|ccc|ccc|ccc}
 &
  \textbf{} &
  \multicolumn{1}{l|}{} &
  \multicolumn{3}{c|}{\textbf{Words Per Sentence}} &
  \multicolumn{3}{c|}{\textbf{Culture}} &
  \multicolumn{3}{c|}{\textbf{politic}} &
  \multicolumn{3}{c|}{Lifestyle} &
  \multicolumn{3}{c}{money} \\ \hline \hline
 &
  \multicolumn{1}{c|}{} &
  \# articles &
  $\mu$ &
  $\sigma$ &
  $d$ &
  $\mu$ &
  $\sigma$ &
  $d$ &
  $\mu$ &
  $\sigma$ &
  $d$ &
  $\mu$ &
  $\sigma$ &
  $d$ &
  $\mu$ &
  $\sigma$ &
  $d$ \\ \hline \hline
\multicolumn{1}{l|}{} &
  \multicolumn{1}{l|}{Top Stories} &
  148 &
  20.64 &
  4.09 &
  0.93 &
  2.27 &
  2.18 &
  0.43 &
  1.37 &
  1.97 &
  0.47 &
  3.60 &
  1.70 &
  0.77 &
  0.44 &
  1.10 &
  0.69 \\
\multicolumn{1}{l|}{} &
  \multicolumn{1}{l|}{Australia} &
  46 &
  21.11 &
  4.18 &
  0.86 &
  2.23 &
  2.02 &
  0.44 &
  1.02 &
  1.59 &
  0.59 &
  2.81 &
  1.71 &
  0.96 &
  0.41 &
  0.70 &
  0.70 \\
\multicolumn{1}{l|}{\multirow{-4}{*}{Far-right}} &
  \multicolumn{1}{l|}{Finance} &
  13 &
  21.56 &
  3.56 &
  0.80 &
  1.48 &
  0.77 &
  0.68 &
  0.79 &
  0.80 &
  0.67 &
  \textcolor{gray!60}{ 9.12} & 
  \textcolor{gray!60}{ 2.44} &
  \textcolor{gray!60}{ --} &
  \textcolor{gray!60}{ 5.30} &
  \textcolor{gray!60}{ 2.44} &
  \textcolor{gray!60}{ --}\\
\multicolumn{1}{l|}{\multirow{-3.5}{*}{publishers}} &
  \multicolumn{1}{l|}{Climate change} &
  12 &
  \textcolor{gray!60}{ 21.60} & 
  \textcolor{gray!60}{ 4.60} &
  \textcolor{gray!60}{ --} &
  \textcolor{gray!60}{ 2.03} &
  \textcolor{gray!60}{ 1.80} &
  \textcolor{gray!60}{ --} &
  \textcolor{gray!60}{ 1.61} &
  \textcolor{gray!60}{ 1.86} &
  \textcolor{gray!60}{ --} &
  \textcolor{gray!60}{ 3.64} &
  \textcolor{gray!60}{ 2.24} &
  \textcolor{gray!60}{ --} &
  \textcolor{gray!60}{ 0.67} &
  \textcolor{gray!60}{ 1.00} &
  \textcolor{gray!60}{ --} \\ \hline
\multicolumn{1}{l|}{} &
  \multicolumn{1}{l|}{Top Stories} &
  9459 &
  23.77 &
  9.96 &
  0.40 &
  1.98 &
  2.49 &
  0.57 &
  1.28 &
  2.09 &
  0.59 &
  4.70 &
  3.18 &
  0.54 &
  1.10 &
  1.99 &
  0.53 \\
\multicolumn{1}{l|}{} &
  \multicolumn{1}{l|}{Australia} &
  6880 &
  23.50 &
  6.90 &
  0.52 &
  2.37 &
  2.61 &
  0.42 &
  1.68 &
  2.29 &
  0.41 &
  5.06 &
  3.60 &
  0.42 &
  1.48 &
  2.14 &
  0.37 \\
\multicolumn{1}{l|}{\multirow{-4}{*}{Moderate}} &
  \multicolumn{1}{l|}{Finance} &
  1073 &
  22.79 &
  7.11 &
  0.62 &
  1.86 &
  1.73 &
  0.58 &
  0.94 &
  1.35 &
  0.66 &
  9.64 &
  3.34 &
  -0.75 &
  6.17 &
  3.30 &
  -1.25 \\
\multicolumn{1}{l|}{\multirow{-4}{*}{publishers}} &
  \multicolumn{1}{l|}{Climate change} &
  881 &
  24.48 &
  8.91 &
  0.37 &
  2.27 &
  2.46 &
  0.43 &
  1.57 &
  2.15 &
  0.43 &
  4.85 &
  2.47 &
  0.46 &
  1.51 &
  1.87 &
  0.33 \\ \hline \hline
\multicolumn{2}{c|}{far-right Twitter shared} &
  11643 &
  27.13 &
  6.98 &
   &
  3.60 &
  3.11 &
   &
  2.78 &
  2.86 &
   &
  6.67 &
  4.02 &
   &
  2.46 &
  2.92 &
  
\end{tabular}}
  \end{table*}

Here, we assess whether there is a difference in linguistic patterns based on content in the articles that far-right \textit{users} \textit{share}.
Note there is a difference with the analysis in the previous \cref{subsec:far-right-publishers} in which we analyzed the difference in information \emph{production} by far-right publishers, here we investigate information \emph{consumption and sharing} by far-right users.

\subsubsection{Setup.}
The far-right Twitter users shared $11,643$ articles from the Google News dataset on the four topics of interest -- `Top Stories,' `Australia,' `Finance,' and `Climate Change'.
We process these articles' content using LIWC.
Finally, we perform \emph{t-tests} (Bonferroni corrected) to compare articles shared by the far-right Twitter users with 1) articles produced by far-right publishers and 2) articles produced by moderate publishers.
\cref{tab:farrighttext} shows select LIWC categories that show statistically significant differences. There are several noteworthy findings which we introduce below.



\subsubsection{Results.}
First, we find $64$ (LIWC category, article topic) pairs statistically significantly different when comparing articles from far-right producers with articles shared by far-right Twitter users.
In comparison, there are $325$ significantly different pairs between articles from moderate publishers and those shared by far-right Twitter users.
However, only $21$ pairs significantly differ between far-right and moderate publishers (see \cref{subsec:far-right-publishers}).
This indicates that the overall output of far-right publishers is generally more similar to moderate publishers in content-based linguistic patterns, however the far-right Twitter users share articles that exhibit different and less common linguistic patterns.

Investigating the mean values ($\mu$) and standard deviation ($\sigma$) yields a similar conclusion -- the differences between what the far-right Twitter users chose to share and what is produced by the far-right publishers are larger than the differences between far-right and moderate publishers.
%
Second, the LIWC categories for the articles that the far-right users shared showed significant differences in the categories \textsc{Words Per Sentence (WPS)}, \textsc{Culture}, and \textsc{politic}, with the articles Twitter shared showing significantly higher mean values. 
The \textsc{Culture} category includes words relating to nations, political processes (\textsc{politic}), and ethnic identities.
\textsc{Lifestyle} and \textsc{money} also showed significant differences in the topics of `Top stories' and `Australia' with high effect sizes. \textsc{Lifestyle} includes words that discuss money, households, employment, and religion. This indicates that far-right users typically shared articles that were about politics, societal make-up, and money -- common key issues for right-aligned groups.
%
Third, the articles shared by Twitter users use more words from the categories of \textsc{Culture} and, specifically, its subcategory, \textsc{politic}, compared to the far-right-leaning articles from the Google news dataset. 
This indicates that far-right Twitter users share more political articles than the general Google News sample suggests.
This seems intuitive, as most conspiracy theories touch on politics.

\subsubsection{Conclusion.}
The above results suggest that the information consumption patterns of online far-right users differ from the scope of article production by far-right publishers.
We hypothesize that these users do not share random samples of the articles produced by far-right publishers.
Rather, they rather cherry pick the articles most useful for their arguments.
We note that users may not necessarily express agreement or a positive point of view of the article that they share.
In fact, far-right Twitter users regularly share articles from reputable and left-leaning publishers, potentially as evidence for a circumstance they wish to condemn. This suggests that far-right users are willing to draw on any news articles that evidence their viewpoints, regardless of the political leaning of the source (although they seem to prefer far-right over moderate produced articles, see \Cref{fig:venn}).
This is quite different from what has been proposed by previous research, which is that users' engagement with a news article is heavily mediated by the source it is attributed to, and whether the user has an overall positive or negative view of the news media company itself~\citep{yun2018hostile}.  

\section{The Style: Style over Content}
\label{styleovercontent}
In the previous sections, we learned that the integrity of contents do not have strong association with the ideologies of the publishers. Additionally, contents produced by far-right publishers and consumed by far-right users are different which was unexpected. In this section, instead of comparing contents, we investigate the style of texts from different groups and distinguish between them.

\subsubsection{Why style over content?}
Style words reflect how people communicate, whereas content words convey what they say in terms of meaning and topic. 
Style words are more closely linked to measures of people's social and psychological worlds~\citep{tausczik2010psychological}. 
Styles encompass a range of linguistic features, including sentence structure, grammar, and punctuation patterns. 
Unlike the content, which continuously changes and can be influenced by subject matter, external sources, or intentional deception, stylistic features are intrinsic to one's writing and are less prone to deliberate manipulation~\citep{da2022text}.
%


\subsection{Identify Extreme Groups using Styles}
\label{subsec:misleadingstyles}
Here, we answer RQ2 by proposing a style classifier and demonstrating that we can distinguish ordinary
online communities from different online extreme groups
based on style alone.

\citet{lee2022whose} showed that political groups with different ideologies exhibit distinct tendencies when consuming and disseminating information on social media platforms. 
In order to investigate styles of ideological online groups and examine if different groups show different text styles,
we use the two collections of Facebook groups that represents two datasets of extreme groups -- antivax and far-right, as detailed in~\cref{dataset}.


\subsubsection{Method and design.}
We begin by demonstrating the effectiveness of using style to distinguish between extreme groups and a ``benign'' control group. 
We use the \textsf{scikit-learn} Python library\footnote{\url{https://scikit-learn.org/stable/modules/multiclass.html}} for all classifiers used in this section;  Logistic regression (LR), Linear SVC (SVC), and Random Forest (RF). 
All three classifiers inherently support multiclass classification, and we use the default strategy for each classifier. 

We design a predictive experiment aimed at evaluating and contrasting the effectiveness of three dictionary-based stylistic metrics (\textsc{LIWC}, \textsc{Grievance} and \textsc{StyloMetrix}).
The task is to distinguish the posts among the three Facebook groups; two extreme Facebook groups and a \textsc{normal} (non-extreme) group. We created the \textsc{normal} group to encompass a wide range of discussions, from cooking to non-profit organizations.
To put the performance of stylistic classifiers into context, we compare against a content-based baseline -- the popular text encoding technique, \textsc{BERT}~\cite{devlin2018bert} commonly used in textual classification tasks. 
In contrast to the three dictionary-based encodings, \textsc{BERT} considers the contextual information for each instance of a given word, enhancing its capabilities. We use \textsf{spacy-sentence-bert} Python library\footnote{\url{https://github.com/MartinoMensio/spacy-sentence-bert}} to produce vectors of given text. The pretrained model of \texttt{en\char`_stsb\char`_distilbert\char`_base} is used which transforms each sentence into a vector of $768$ dimensions. \textsf{spacy-sentence-bert} is a wrapper of \textsf{sentence-BERT} package which uses mean-pooling strategy by default~\citep{reimers2019sentencebert}. 	
This approach captures contextual information from the entire input sentence and produces fixed-length vectors suitable for downstream tasks. 


\begin{table}[]
\caption{
  Performance of five feature sets.
  We distinguish between two extreme groups -- \texttt{Far right} and \texttt{Antivax} -- and the \textsc{normal} Facebook group. 
  We randomly sample $1,000$ posts for each group.
  We report accuracy and macro-F1 to compare among the classifiers and features used. We boldfaced the best results using \textsc{BERT} and \textsc{LGS}. 
}
\label{classification}
\large
  \centering
  \resizebox{0.48\textwidth}{!}{
    \begin{tabular}{lc|c|ccc|c}
 &  & \multicolumn{1}{l|}{BERT} & \multicolumn{1}{l}{\textsc{LIWC}} & \multicolumn{1}{l}{\textsc{Grievance}} & \multicolumn{1}{l|}{\textsc{StyloMetrix}} & LGS \\ \hline \hline
\multirow{2}{*}{LR} & accuracy & \bf{0.80}  & 0.73 & 0.56 & 0.69 & 0.74 \\
                                     & macro F1 & \bf{0.79} & 0.73 & 0.54 & 0.68 & 0.74 \\ \hline
\multirow{2}{*}{SVC}          & accuracy & 0.76 & 0.66 & 0.57 & 0.71 & 0.68 \\
                                     & macro F1 & 0.76 & 0.61 & 0.55 & 0.70 & 0.64 \\ \hline
\multirow{2}{*}{RF}       & accuracy & 0.75 & 0.72 & 0.65 & 0.70 & \textbf{0.77} \\
                                     & macro F1 & 0.75 & 0.71 & 0.63 & 0.69 & \textbf{0.76} \\ \hline
    \end{tabular}
    }
\end{table}

This exercise aims to evaluate various feature sets, not the classification algorithms.
We interpret the difference in prediction performance as a difference in the representativity of the feature sets.
\cref{classification} reports classification performance of three off-the-shelf classifiers.
We randomly sampled $1,000$ posts from each list of groups (\texttt{Far right}, \texttt{Antivax} and \textsc{normal}). 
The reported performance is the average result of 5-fold cross-validation.
We compare the three basic stylistic feature sets (\textsc{LIWC}, \textsc{Grievance} and \textsc{StyloMetrix}) and their concatenation -- denoted as \textsc{LGS}.

\subsubsection{Results.}
Since BERT leverages the content and style of the given text, it is often the best-performing feature set.
However, some stylistic-based classifiers outperform the content-based classifiers: the Random Forest classifier, \textsc{LGS}, outperforms BERT. 
In other words, based solely on the style and without leveraging the content of a given post, we can predict which ideological group the post came from.
The true positive rate for the \textsc{normal} group was $0.95$, $0.79$ for the \texttt{Far right} group, and $0.66$ for the \texttt{Antivax} group.
We notice that some posts from \texttt{Antivax} group were misclassified as \texttt{Far right}, while only a few posts from \texttt{Far right} were misclassified as \texttt{Antivax}. 
This is because the anti-vax space is a unique niche with far-right elements, while far-right groups generally discuss diverse topics. 

\subsection{Identifying Styles}
\label{subsec:identify-styles}
Here, we identify and classify the writing styles of posts used by people in fringe Facebook communities.
This predictive exercise differs from the one in \cref{subsec:misleadingstyles};
here, we test whether human-labeled styles can be detected using a style-based classifier;
we distinguished user groups in \cref{subsec:misleadingstyles}.

\subsubsection{Method and design.}
First, a team member with a writing background manually annotated the writing style of $100$ text samples from the \texttt{Far right} and \texttt{Antivax} groups ($50$ for each group).
She employed a deep qualitative approach in which the style labels were generated alongside the labeling.
We identified styles that are extensively used in misinformation online environments by manually inspecting them. The labeled styles in these samples are ``Casual'', ``Empowerment'', ``Clickbait'', ``Expert'' and ``Intimacy''. 
This annotation is a result of sampling 50 posts from both \texttt{Far right} and \texttt{Antivax} then we remove the posts with no style; \cref{tab:style} shows the number of posts per style.

\begin{table}[]
\caption{Number of Facebook posts per style.}
\label{tab:style}
\large
  \centering
  \resizebox{0.48\textwidth}{!}{
\begin{tabular}{c|ccccc|c}
          & Casual & Empowerment & Clickbait & Expert & Intimacy & Total \\ \hline \hline
\textsc{Far right} & 11     & 21          & 1         & 13     & 1        & 47    \\
\textsc{Antivax}   & 11     & 4           & 1         & 0      & 0        & 16    \\ \hline
     & 22     & 25          & 2         & 13     & 1        & 63  
\end{tabular}
}
\end{table}

\begin{table}[]
\caption{Classification results of the three styles; ``Casual'', ``Empowerment'' and ``Expert''. A Random Forest (RF) classifier and a Logistic Regression (LR) was used with a stratified split. The results show the average performance of 2-fold cross-validation due to the size of the samples.}
\label{styles}
\large
\centering
\resizebox{0.48\textwidth}{!}{
\begin{tabular}{l|c|rr|rr}
\multicolumn{1}{c|}{\multirow{2}{*}{Style}} &
  \multicolumn{1}{c|}{\multirow{2}{*}{Classifier}} &
  \multicolumn{2}{c|}{LGS} &
  \multicolumn{2}{c}{BERT} \\ \cline{3-6} 
\multicolumn{1}{c|}{} &
  \multicolumn{1}{c|}{} &
  \multicolumn{1}{c}{macro F1} &
  \multicolumn{1}{c|}{Accuracy} &
  \multicolumn{1}{c}{macro F1} &
  \multicolumn{1}{c}{Accuracy} \\ \hline \hline
\multirow{2}{*}{Empowerment} & RF       & \textbf{0.60} & \textbf{0.58} & 0.51          & 0.54          \\
                             & LR & 0.49          & 0.52          & \textbf{0.67} & \textbf{0.66} \\ \hline
\multirow{2}{*}{Casual}      & RF       & \textbf{0.67} & 0.5           & 0.54          & \textbf{0.55} \\
                             & LR & 0.45          & 0.52          & \textbf{0.67} & \textbf{0.70} \\ \hline
\multirow{2}{*}{Expert}      & RF       & \textbf{0.47} & \textbf{0.5}  & 0.42          & \textbf{0.5}  \\
                             & LR & 0.40          & \textbf{0.73} & \textbf{0.57} & 0.46          \\ \hline
\end{tabular}}
\end{table}

``Clickbait'' and ``Intimacy'' have only two and one exemplars, respectively, and we decided to remove them from the rest of this analysis.
As a result, we classify solely the styles ``Casual'', ``Empowerment'' and ``Expert''. 
We performed a binary classification for each style using One-Versus-Rest (OvR) strategy by randomly sampling negative samples to balance the sample size.  
For the Random Forest classifier, we used max tree depth = $3$ and the number of trees = $8$ due to the small sample size. 
\subsubsection{Results.}
\cref{styles} reports the classification results. 
In general, the stylistic features (\textsc{LGS}) perform better with the Random Forest classifier; \textsc{BERT} shows better performance with the Logistic Regression. This is expected since in \textsc{BERT} embeddings, each dimension is not inherently interpretable or separable and thus not ideal for the Random Forest classifier which excels with features that individually carry strong predictive signals. 
While it is true that the  Random Forest classifier may not fully leverage the rich contextual information in BERT embeddings, we aimed to assess whether it could still yield valuable insights or competitive results when compared to more complex models. This experimental choice was made to offer a comparative perspective on different feature representations and classifiers.
  
For all three styles, the \textsc{LGS} stylistic features outperform \textsc{BERT} with the Random Forest classifier measured by macro F1 score, which is consistent with the results presented in~\cref{classification} for group style classification.
``Expert'' style was the most challenging for both classifiers since there were only $13$ labeled samples.

\subsection{Identify Far-right articles using Styles: production vs. consumption}
\label{pro_con_classify}
Here, we answer RQ3 by distinguishing production from sharing. 
Inspired by the findings in~\cref{pvss}, we hypothesise that far-right production can be differentiated from far-right consumption on the basis of style. We train the style classifier to distinguish far-right production from far-right consumption, and add moderate production for comparison purposes. 

\subsubsection{Method and design.}
We use the $155,669$ articles shared by far-right Twitter users (see \cref{dataset}) as far-right consumption data (FRCons) since far-right Twitter users shared these. 
In order to increase the number of articles produced by far-right publishers in our sample, as well as the number of moderate articles, we expand the number of articles as follows. Each article shared by Twitter users is added to either the moderate or far-right category based on the leaning of the publisher, as long as the article's publisher is referenced in {\it allsides} data. After this expansion, we have $155,669$ Twitter articles shared by far-right users on Twitter, (FRCons), $41,995$ articles from moderate producers (MDProd), and $14,758$ far-right produced articles (FRProd). 

\begin{figure}[tb]
  \includegraphics[width=0.32\textwidth]{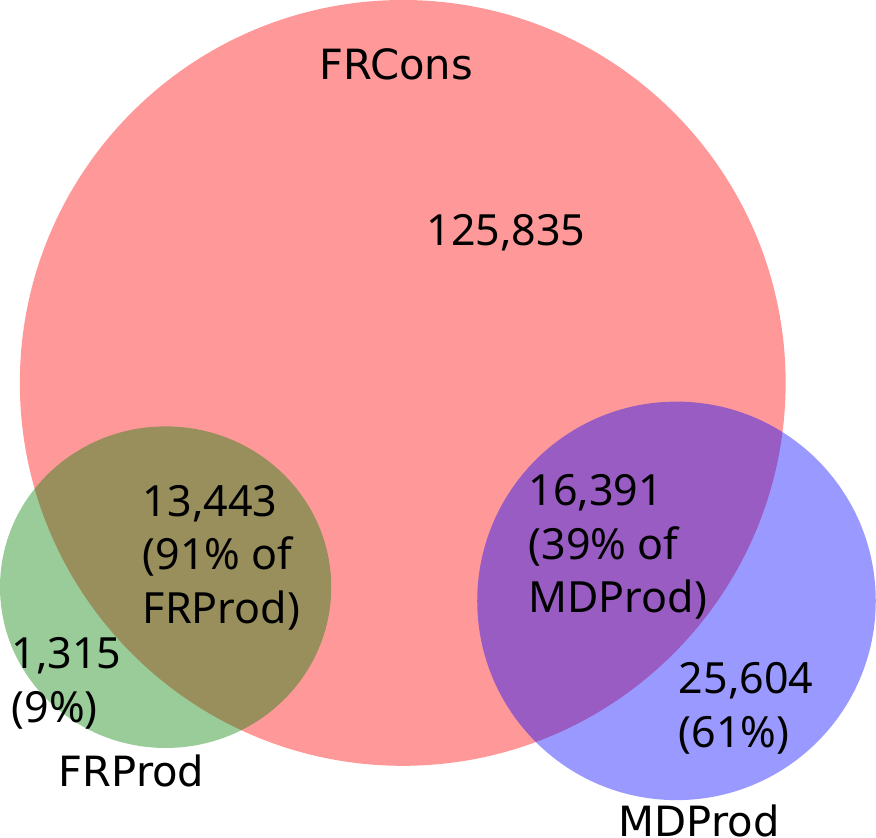}
  \centering
  \caption{Venn diagram showing the intersection of articles produced (FRProd, MDProd) and consumed (FRCons).}
  \label{fig:venn}
\end{figure}

\Cref{fig:venn} shows the intersection size between these three sets of articles as a Venn diagram.
$91\%$ of all far-right produced articles are shared by the far-right consumers, whereas only $39\%$ of the moderate articles are shared by the far-right users.
Proportionally, the intersection between far-right produced (FRProd) and far-right shared (FRCons) is larger than the intersection between moderate-produced (MDProd) and far-right shared (FRCons). 
This shows that, while far-right users opportunistically link to articles from both far-right and moderate producers, they prefer far-right producers.

We build a textual classifier that uses the text of each article and predicts whether it is far-right produced, moderate produced or far-right shared.
We downsample each class to the smallest class size ($14,758$ articles).

\subsubsection{Results.}
In \cref{subsec:identify-styles}, we observed that our curated stylistic features (\textsc{LGS}) perform better with the Random Forest classifier while \textsc{BERT} features outperform with Logistic Regression in most cases for identifying styles.

\begin{figure}[tbp]
  \includegraphics[width=.4\textwidth]{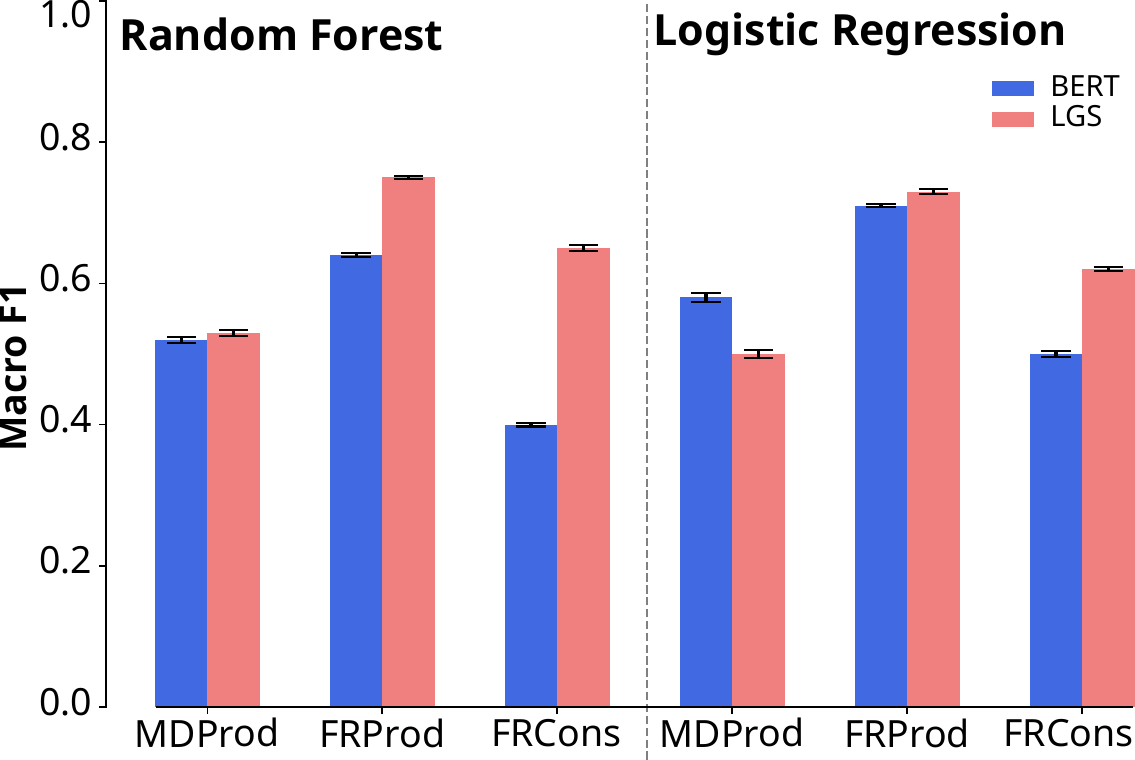}
  \centering
  \caption{Classification results of the moderate-produced articles (MDProd), far-right produced articles (FRProd) and the articles consumed by far-right users (FRCons). A Random Forest classifier and a Logistic Regression was used to report macro F1-score from 10-fold stratified cross-validation. The error bars represent 95\% confidence intervals. We use `lbfgs' solver with `multi\_class' option set to `multinomial' to support three class classification for Logistic Regression.}
  \label{consumption_results}
\end{figure}

In \Cref{consumption_results}, we see that \textsc{LGS} with the Random Forest classifier consistently outperforms \textsc{BERT} when distinguishing production from consumption. In addition, \textsc{LGS} with Logistic Regression, exempting for articles from moderate publishers, outperforms \textsc{BERT} as well. Particularly, \textsc{LGS} outperforms \textsc{BERT} by far when distinguishing the far-right consumption group (FRCons). 
This indicates that it is the style, not the content, of the articles that better characterizes the far-right consumption patterns. 

\begin{figure}[]
\includegraphics[scale=0.39]{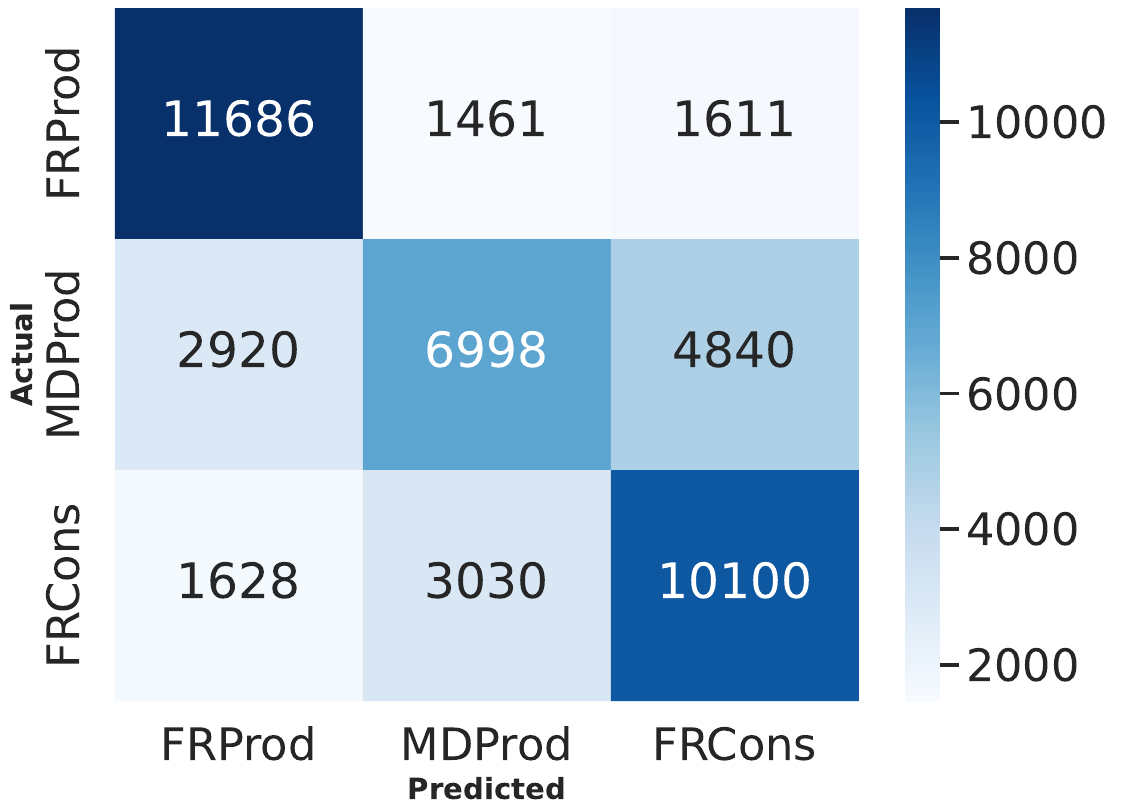}
\centering
\caption{Confusion matrix of the three classes from Random Forest classifier using \textsc{LGS} features. }
\label{confusion}
\end{figure}

\Cref{confusion} reports the confusion matrix of the three class classification from the Random Forest classifier. This result is the average of 10-fold stratified cross-validation. The most notable observation is that far-right-produced articles (FRProd) are easily distinguished from the others and far-right consumption (FRCons). The most confusions occurred between the far-right consumption (FRCons) and the moderate produced (MDProd). This result indicates that while far-right production (FRProd) and consumption (FRCons) are clearly separable, the styles utilized by moderate production (MDProd) and far-right consumption (FRCons) are similar. Also, since we did not find a significant difference between far-right and moderate articles in terms of their content (\cref{subsec:far-right-publishers}), it is implied that the styles are better signals for information consumers. 

These findings reinforce the previous observations that far-right production and far-right consumption patterns are different. More specifically, far-right users do not exclusively consume far-right content but rather cherry pick articles that pursue certain styles. Indeed, style could be a contributing factor driving political polarisation in and of itself. Most notably, the findings imply that the distinctiveness of far-right texts compared to moderate sources is their style; and from this, it follows that perhaps a more distinctive approach to styling moderate views, tailored to different audiences, could help moderate perspectives compete against fringe views in today's attention economy.   

\section{Related Work}
In this section, we investigate content-based misinformation detection methods, their associated limitations, and style-based approaches.

\subsubsection{Content-based methods.}
Traditionally, misinformation detection techniques have relied on content-based information, such as encoded texts using language models. 
\citet{alkhodair2020detecting} proposed a recurrent neural network model for detecting rumors by utilizing Word2Vec~\citep{mikolov2013efficient} representation. They showed that their model outperforms the state-of-the-art sequential classifier from~\citet{zubiaga2016learning}. They then applied their model to emerging breaking news in a real-time Twitter stream. The F1 scores for the two case studies were $0.757$ and $0.791$. Our style-based classifier achieved comparable results when distinguishing far-right production and far-right consumption.  
Similarly, \citet{horne2017just} integrated three types of features, including stylistic elements, to classify fake news. 
Their findings revealed significant distinctions between fake and real news content. 
Meanwhile, \citet{sarnovsky2022fake} concentrated on identifying fake news within the Slovak online sphere. 
They applied deep learning models to Word2Vec representations of the texts. 
While the models demonstrated remarkable performance (best model reaching the accuracy of $98.93\%$), it is worth noting that the dataset predominantly comprised articles centered around COVID-19.

A  shortcoming of content-based methods lies in the dynamic nature of misinformation topics. 
Models trained on predefined subjects may struggle to adapt to emerging themes. 
Furthermore, training deep-learning models requires substantial data to mitigate the risk of over-fitting. 
To address these limitations, \citet{raza2022fake} introduced a context-based model focusing on social aspects to identify fake news. 
This model incorporates users' social interactions, such as comments on news articles, posts, and replies, as well as upvotes and downvotes. 
This approach can serve as a valuable complement, especially when engagement data is readily available.

\subsubsection{Style-based methods.}
\citet{whitehouse2022evaluation} showed that general-purpose content-based classifiers tend to overfit to specific text sources.
In response, the authors proposed a style-based classification approach. The proposed stylometric features, however, leverage the categories in the General Inquirer (GI) dictionary, encompassing content-specific words related to religion and politics. To factor out content as much as possible, we omit such specific categories when employing the Linguistic Inquiry and Word Count (LIWC).

More recently, \citet{kumarage2023stylometric} demonstrated the effective detection of AI-generated texts in Twitter timelines using various stylistic signals. Their study showed that classifiers employing the proposed stylometric features outperformed Bag of Words~\citep{manning2008introduction} and Word2Vec embeddings~\citep{mikolov2013efficient}. Notably, they mentioned that among the stylometric features, punctuation and phraseology features proved to be the most significant. While these findings are motivating, their research primarily focuses on distinguishing between human and AI-authored content within a given Twitter timeline. In contrast, we aim to discriminate between different writing styles, such as intimate and expert styles. 

In another attempt to verify authorship using style representations content-controlled style representations have been proposed~\citep{wegmann-etal-2022-author}. The authors demonstrated that performance varies when controlling for different levels of contents, e.g., authorship verification within texts from the same conversation or the same domain. They used a clustering algorithm on text samples and manually inspected the resulting clusters of texts to find out what styles were learned, such as `punctuation'. In our proposed work, we attempt to learn and distinguish styles used by extreme groups since we observed that these groups strategically adopted certain styles to reach vulnerable demographics in online misinformation space. This is different from learning innate styles of individuals which maybe more subtle.              


\section{Conclusion and Discussion}
We have shown that misinformation is conveyed through styled messages by detecting styles in extreme online communities and being able to distinguish the communities using styles rather than content. Specifically, Facebook pages that share misinformation can be distinguished by the style of the posts made within each group. The classification results showed that stylistic measures can outperform the content-based classifier. This is intriguing since the two groups, especially an anti-vaccination group, can be easily identified by analyzing content-related vocabulary.

We also evidenced that content produced by the far-right differs significantly from far-right online users' consumption. This is to say that while producers may cover a wide range of topics, users selected a narrow set of articles to share, which had consistent style features. Thus, we showed that the far-right production and the far-right consumption can be distinguished by using stylistic features. This indicates that misinformation consumers prefer certain styles of information as opposed to the contents of information. 

This study has limitations across multiple dimensions. Firstly, the Facebook group dataset is constrained in its post volume. Additionally, the availability of labeled style data was even more scarce, limiting the generalizability of the findings. Our plan in the future is to apply an active learning process to label more posts efficiently. Secondly, we focused only on the styles of messages in order to distinguish and observe the effects of styled text. While studying online misinformation, we noticed that misinformation producers strategically adopt styles to effectively spread the information and utilize formats such as graphics, videos and reports. We can enhance the results by considering these facets of misinformation packaging.

Lastly, we intend to explore and develop our understanding of the vulnerable demographics in online spaces. We observed that extremist online groups aimed at these personas tailor their content to these preferences. A better understanding of the patterns and styles of misinformation the vulnerable demographics are attracted to can serve as a guiding reference for policymakers, suggesting adopting a comparable strategy to effectively counter misinformation dissemination. 

\section*{Ethical Statement}
\subsubsection{Data Collection and Management}
All data that we obtained was publicly available at the time of data collection. We discarded deleted, protected, and redacted 
content at the time of analysis. Therefore, the analyses reported in this work does not compromise any user privacy. This project was approved by the Human Ethics Committee of our institution (approval number redacted for review).



\subsubsection{Broader Potential Impact of Work}
By exploring the landscape of Australian news media and the online misinformation space, we hope to raise awareness of how language styles used in online communities may appeal to vulnerable demographics. Understanding how misinformation is styled and packaged provides insights into how we can approach affected populations with reliable information.  
Awareness goes both ways, however, and the results presented in the paper could also lead to malicious users learning ways to easily manipulate a target audience. Based on our observations, however, creators of misinformation with the intention to reach a target population are already styling their messages accordingly. By showing that misinformation consumers are attracted to styled messages, we hope any entities involved in misinformation intervention campaigns can utilize these findings. 

\bibliography{main}

\newpage

\clearpage
\appendix
\onecolumn
\normalsize

Supplementary materials accompanying the paper \textit{Misinformation is not about Bad Facts: An Analysis of the Production and
Consumption of Fringe Content}.

\section{Details of Stylistic measures}
\label{liwc_list}
We summarize the linguistic measures and list the features that we used in~\cref{styleovercontent}. The full list of features of \textsc{StyloMetrix} is found here\footnote{\url{https://github.com/ZILiAT-NASK/StyloMetrix/blob/v0.1.0/resources/metrics_list_en.md}}.

\begin{table*}[h]
 \caption{Summary of the linguistic tools used in~\cref{styleovercontent}. }
  \label{dictionary}
  \centering
  \large
  \resizebox{0.95\textwidth}{!}{
\begin{tabularx}{\textwidth}{c|>{\centering\arraybackslash}X|>{\centering\arraybackslash}X|>{\centering\arraybackslash}X}
  & \textsc{LIWC} & \textsc{Grievance} & \textsc{StyloMetrix} \\ \hline \hline
  
Summary &
LIWC is a transparent text analysis program that counts words in psychologically meaningful categories &
  Grievance dictionary assess grievance-fuelled communications through language &
 StyloMetrix is a tool for creating text representations as StyloMetrix vectors \\ \hline
\# features &
  89 &
  22 &
  175 \\ \hline
  features & Segment, WC, Analytic, Clout, Authentic, Tone, WPS, BigWords, Dic, Linguistic, function, pronoun, ppron, i, we, you, shehe, they, ipron, det, article, number, prep, auxverb, adverb, conj, negate, verb, adj, quantity, Drives, affiliation, achieve, power, Cognition, allnone, cogproc, insight, cause, discrep, tentat, certitude, differ, memory, Affect, tone\_pos, tone\_neg, emotion, emo\_pos, emo\_neg, emo\_anx, emo\_anger, emo\_sad, swear, Social, socbehav, prosocial, polite, conflict, moral, comm, socrefs, substances, risk, curiosity, allure, Perception, attention, motion, space, visual, auditory, feeling, time, focuspast, focuspresent, focusfuture, Conversation, netspeak, assent, nonflu, filler, AllPunc, Period, Comma, QMark, Exclam, Apostro, OtherP &  deadline, desperation, fixation, frustration, god, grievance, hate, help, honour, impostor, jealousy, loneliness, murder, paranoia, planning, relationship, soldier, suicide, surveillance, threat, violence, weaponry & \\ \hline
  
\end{tabularx}}
\end{table*}


\end{document}